%% file: main.tex
\documentclass[10pt, a4paper, twocolumn, teaser, showabstract]{naverlabseurope}

\usepackage{CJKutf8}
\usepackage{lipsum}
\usepackage{multicol}
\usepackage{tikz,tkz-kiviat,pgfplots}

\usepackage{pifont}
\usepackage{multirow}

\newcommand{\ours}{NAVER LABS Europe Submission to the Instruction-following Track}
\newcommand{\fix}[1]{\textcolor{red}{ #1}}
\newcommand{\seamless}{\texttt{SeamlessM4T-v2-large}}
\newcommand{\llama}{\texttt{Llama-3.1-8B-Instruct}}

\graphicspath{{figures/}}

\title{NAVER LABS Europe Submission to the Instruction-following Track}
\titlerunning{\ours}

\authors{Beomseok Lee$^{123\star}$, Marcely Zanon Boito$^{1\star}$, Laurent Besacier$^{1}$, Ioan Calapodescu$^{1}$}
\affiliations{$^1$NAVER LABS Europe, France\\
$^2$University of Trento, Italy\\$^3$Fondazione Bruno Kessler, Italy}
\correspondingauthor{marcely.zanon-boito@naverlabs.com}
\contributions{$^{\star}$equal contribution}

\teaserfig{\fbox{\includegraphics[width=0.95\textwidth]{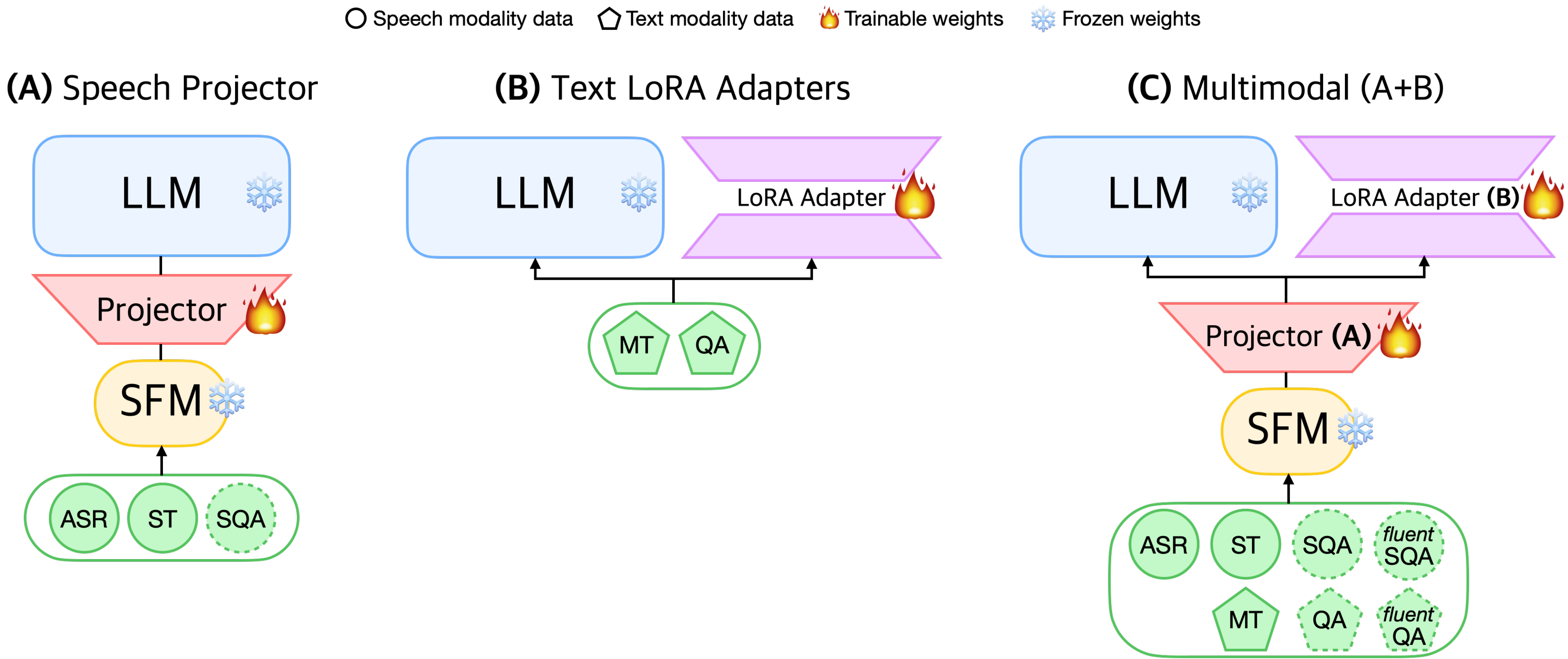}}}

\teasercaption{Our training pipeline. A speech projector~(A) and text LoRA adapters~(B) are trained in parallel using speech-to-text and text-to-text data, respectively. These modules are then integrated during a brief multimodal adaptation step~(C).\label{fig_apx:main_figure}}

\input{paper/0_abstract}

\begin{document}

\maketitle

\input{paper/1_introduction_section}

\input{paper/2_data_section}

\input{paper/3_architecture_section}

\input{paper/4_results_section}
\input{paper/5_submission_section}
\input{paper/6_conclusion}

\section*{Acknowledgments}

We thank our NLE colleagues Vassilina Nikoulina, for the implementation of the LLM-as-judge scripts using \texttt{bergen}, and Christian Wolf, for kindly verifying some of our German ST/SQA outputs. 
This work was partially funded by the European Horizon 2022 project UTTER (Unified Transcription and Translation for Extended Reality), under grant agreement No 101070631.

{
    \small
    \bibliographystyle{ieeenat_fullname}
    \bibliography{custom}
}

\clearpage
\appendix
\input{supplementary/appendix_section}

\end{document}

%% file: paper/0_abstract.tex
\begin{abstract}

In this paper we describe NAVER LABS Europe submission to the instruction-following speech processing short track at IWSLT 2025. We participate in the constrained settings, developing systems that can simultaneously perform ASR, ST, and SQA tasks from English speech input into the following target languages: Chinese, Italian, and German. Our solution leverages two pretrained modules: (1) a speech-to-LLM embedding projector trained using representations from the \seamless{} speech encoder; and (2) LoRA adapters trained on text data on top of \llama{}. These modules are jointly loaded and further instruction-tuned for 1K steps on multilingual and multimodal data to form our final system submitted for evaluation.

\end{abstract}

%% file: paper/1_introduction_section.tex
\section{Introduction}

Large language models (LLMs) have demonstrated remarkable success across various text-based natural language processing tasks~\cite{achiam2023gpt, touvron2023llama, jiang2024mixtral, yang2024qwen2, alves2024tower,martins2024eurollm}, motivating research into extending them to other modalities. This has led to the development of multimodal LLMs capable of processing speech, audio, images and video~\cite{team2023gemini, driess2023palm, rubenstein2023audiopalm,liu_visual_2023, tang2023salmonn,defossez2024moshi,hu2024wavllm, laurencon_what_2024, huang2024audiogpt, nguyen2024spirit,ambilduke2025tower}.

This year \textit{IWSLT Instruction-following Speech Processing Track} focuses on the leveraging of LLMs and speech foundation models~(SFM) to build solutions capable to perform multilingual tasks from English speech input and textual multilingual instructions~\cite{abdulmumin-etal-2025-findings}. NAVER LABS Europe~(NLE) participates in the constrained setting of the \textit{short track}, where the tasks proposed are automatic speech recognition~(ASR), speech translation~(ST) and multilingual spoken question answering~(SQA). The target languages for ST and multilingual SQA are Chinese, Italian and German. 
The participants are allowed to use the speech backbone \seamless{}~\cite{barrault2023seamlessm4t} and the text LLM \llama{}~\cite{grattafiori2024llama} for both training and data generation.

Our submitted systems leverage all the available data from the constrained settings, together with data automatically obtained using both backbones. 
We train two types of systems in parallel: (1)~speech-to-text ASR/ST/SQA projectors that project the averaged speech representation from the \seamless{} encoder to the embedding space of a frozen \llama{}; (2)~text-only LoRA adapters~\cite{hu2022lora}, plugged on top of the same frozen LLM. Once both systems are separately trained, we show that we can merge them, increasing overall speech performance, by fine-tuning for only 1K steps on multimodal multilingual data.

This system paper is organized as follows. 
Section~\ref{sec:data} describes the preprocessing applied to the data used in this challenge. 
Sections~\ref{sec:models} and~\ref{sec:settings} describe our training pipeline and experimental settings, respectively.
Section~\ref{sec:experiments} presents our experiments and discussion.
Section~\ref{sec:submission} presents the submitted system. Section~\ref{sec:conclusion} concludes the paper.

%% file: paper/2_data_section.tex
\section{Data}\label{sec:data}

For training our models, we leverage the data from the constrained setting: CoVoST2~\cite{wang2020covost}, EuroParlST~\cite{jairsan2020a} and SpokenSQuAD~\cite{lee2018spoken}. 
With the agreement of the organizers, we also take advantage of the \seamless{} to produce extra synthetic speech data~(Seamless TTS) and multilingual text data~(Seamless MT). \llama{} is used to rephrase SQA answers.
ACL~60-60~\cite{salesky-etal-2023-evaluating} is used for validation and evaluation only.
We now present our data preprocessing~(Section~\ref{sec:datapreproc}) and prompt format~(Section~\ref{sec:promptformat}).

\subsection{Data Preprocessing}\label{sec:datapreproc}

We produce both speech-to-text and text-to-text instructions to train our systems. For the aforementioned datasets, we produce the following splits, where * denotes synthetic splits obtained via \seamless{} MT; $\dag$ indicates splits generated with \seamless{} TTS; and $\ddag$ marks those derived through \llama{}-based rephrasing.
\begin{itemize}
    \item \textbf{CoVoST2:}  \begin{itemize}
        \item \small{\textbf{ASR}
        and \textbf{ST/MT} (en-de; en-zh)}
    \end{itemize}
    \item \textbf{EuroParlST:} \begin{itemize}
        \item \small{\textbf{ASR}
        and \textbf{ST/MT} (en-de; en-it)}
    \end{itemize}
    \item \textbf{SpokenSQuAD:} 
    \begin{itemize}
        \item \small{\textbf{ASR}$^\dag$ and 
        \textbf{MT} (en$^\dag$-de*; en$^\dag$-it*; en$^\dag$-zh*)}
        \item \small{\textbf{SQA/QA} (en$^\dag$-en; en$^\dag$-de*; en$^\dag$-it*; en$^\dag$-zh*)}
        \item \small{\textbf{\textit{fluent} SQA/QA} \\(en$^\dag$-en$^\ddag$; en$^\dag$-de$^{*,\ddag}$; en$^\dag$-it$^{*,\ddag}$; en$^\dag$-zh$^{*,\ddag}$)}
    \end{itemize}
    \item \textbf{ACL 60-60:}  \begin{itemize}
        \item \small{ \textbf{ASR} and \textbf{ST/MT} (en-de; en-it*; en-zh)}
    \end{itemize}
\end{itemize}
Below we detail dataset-specific preprocessing. Statistics are presented in Table~\ref{tab:trainingdata}.

\paragraph{CoVoST2 and EuroParlST} CoVoST2 covers English to German and Simplified Chinese language directions. EuroParlST covers English to German and Italian.
ASR splits for these datasets were built by merging the existing language splits and deduplicating the audio files. For both, language-specific ST and MT splits are created by aligning translations to English speech and reference transcriptions, respectively.

\paragraph{SpokenSQuAD} 
The SpokenSQuAD dataset is organized into two jsons, train and test. Each split is organized in themes, each with several paragraphs. For each paragraph, TTS audio files are available, aligned at the sentence level.\footnote{The format is themeID$\_$paragraphID$\_$sentenceID.} For each sentence, questions (each with several answers) are available. We performed the following modifications to this dataset:
\begin{itemize}
    \item \textbf{Duplicated answers:} we removed duplicated answers to the same question using exact string matching, as well as any questions that required more than one audio file to answer, since we are participating in the SHORT track.
    \item \textbf{Validation set:} we created a validation set by selecting the first 20 themes of the training set and removing them from training~(3,102 entries).
    \item \textbf{New TTS audio:} we generate new TTS audio files for the training set using \seamless{}. 
    Resynthesizing the source audio for this dataset was necessary due to existing dataset misalignment, which we detail in Appendix Section~\ref{app:spokensquadtts}.
    \item \textbf{Multilingual SQA/QA:} we created multilingual SQA/QA sets by first translating questions and answers to target languages using \seamless{}. We then use reference-free COMET\footnote{\texttt{Unbabel/wmt22-cometkiwi-da}}~\cite{rei-etal-2022-comet} to filter out all pairs of questions and answers that do not both score at least 0.85. 
    \item \textbf{Invalid splits:} we created the \textit{invalid} SQA sets by deliberately mismatching context and question themes, thereby creating unanswerable examples. The corresponding answers were labeled as “Not answerable” in four languages (English, Italian, German and Chinese), following the guideline answer provided by the task organizers. While we acknowledge that a small, unknown subset of these reassigned questions may still be answerable, we hypothesize that ensuring a thematic mismatch between the reference context and the question is the most effective strategy for minimizing this issue.
    \item \textbf{\textit{Fluent} SQA/QA version:} we created an alternative SQA/QA training set using \llama{} to regenerate the dataset original answers as fluent sentences. The motivation behind this was the observation that, since the original answers are made of an exact extract of the reference audio/text, the model had a difficult time answering some out-of-domain questions fluently.\footnote{In the official IWSLT 2025 test set we observed examples as the following. Question: \textit{``What are the names of the speakers?''} Our model's answer: \textit{``yin and my colleague jiang''}. While the model answer is an exact and correct extraction from the audio, we were unsure about how this would be considered during evaluation.} More details are presented in Appendix Section~\ref{app:spokensquadnewanswers}.
\end{itemize}

\paragraph{ACL 60-60} We use \seamless{} to generate Italian translations, since the data shared only contained en-de and en-zh splits. We leverage the \textit{dev} set for checkpoint selection during training. The \textit{eval} set is used for testing.

\input{tables/dataset_statistics}

\subsection{Prompt Format}\label{sec:promptformat}
The goal of the short track of this challenge is to produce a model that is capable to 1) transcribe English speech; 2) translate English speech into Italian, German and Chinese; 3) Answer multilingual questions using English speech as input. In this setting, the language of the question must match the language of the answer. 

To develop a model capable of smoothly switching between different tasks, we designed task prompts with a consistent structure: regardless of the task (ASR, ST, or SQA), the user turn begins by encapsulating the speech embeddings within textual tags. This is followed on a new line by a task-specific instruction formulated as a question, and finally, another line containing a common suffix. The list of templates used is available in Appendix Table ~\ref{tab:prompts}.

%% file: tables/dataset_statistics.tex
\begin{table}
\centering
\resizebox{\columnwidth}{!}{
\begin{tabular}{c|ccc}\toprule
\multicolumn{1}{c}{\textbf{Dataset}}  & \textbf{Task}                    & \textbf{Language} & \multicolumn{1}{l}{\textbf{\# Samples}} \\\midrule
\multirow{3}{*}{\textbf{CoVoST2}}     
& \textbf{ASR}                     
& en                
& 289,413\\ 
\cline{2-4} 
& \multirow{2}{*}{\textbf{ST/MT}}
& en-de
& 289,413\\
&
& en-zh
& 289,413\\\midrule
\multirow{3}{*}{\textbf{EuroParlST}}
& \textbf{ASR}
& en
& 35,372
\\\cline{2-4} 
& \multirow{2}{*}{\textbf{ST/MT}}
& en-de
& 32,628\\
&
& en-it
& 29,552
\\\midrule
\multirow{12}{*}{\textbf{SpokenSQuAD}} 
& \textbf{ASR}
& en
& 34,003\\\cline{2-4} 
& \multirow{3}{*}{\textbf{MT}}
& en-de
& 39,362\\
&       
& en-it
& 55,030\\
&       
& en-zh
& 25,078\\\cline{2-4} 
& \multirow{4}{*}{\textbf{SQA/QA}} 
& en-en
& $34,003^{\dag}$
\\
&
& en-de
& $6,574^{\dag}$
\\
&             
& en-it
& $16,767^{\dag}$
\\
&       
& en-zh  
& $7,093^{\dag}$ 
\\\cline{2-4}
& \multirow{4}{*}{\textbf{\textit{fluent} SQA/QA}} 
& en-en
& 32,320                                \\
& 
& en-de
& 4,169                                   \\
&   
& en-it
& 13,712                                  \\
&              
& en-zh
& 3,424      \\
\bottomrule                            
\end{tabular}}
\caption{Training sets statistics by task. For ST/MT sets, target side is duplicated. For SpokenSQuAD, \textdagger{} highlights that the source speech is used twice~(valid and invalid questions, as described in Section \ref{sec:datapreproc}).}
\label{tab:trainingdata}
\end{table}

%% file: paper/3_architecture_section.tex
\section{Training Pipeline}\label{sec:models}

Our training pipeline is illustrated in Figure~\ref{fig_apx:main_figure}. We first train a speech projector on speech tasks~(A), and text LoRA weights on textual tasks~(B). These modules are then reloaded and adapted together on both speech and textual tasks~(C). In this section we describe the key components used and the data sampling strategy.

\paragraph{Foundation Models} For speech, we leverage \seamless{}  model, extracting speech representations for all our audio data from its 24th speech encoder layer~(i.e. the last layer). Prior to training, we average every 3 consecutive frame vectors, reducing significantly the sequence length. This simple trick allows us to train our models with larger batches, while maintaining good performance in speech tasks. All our models are built on top of a frozen \llama{}.

\paragraph{Speech Projector Architecture} 
The speech projector consists of 4 Transformer encoder layers, each with 8 attention heads. The input dimension is set to 1,024, the feed-forward network dimension to 2,048, and the output dimension to 4,096 to align with the embedding size of \llama{}. A dropout rate of 0.1 is applied throughout, and the model employs pre-layer normalization.

\paragraph{LoRA Adapters} LoRA adaptation~\cite{hu2022lora} is applied to both the self-attention (Q/K values, output projection) and feed-forward modules, and across all LLM layers. We use $rank=8,   \alpha=16$. We do not use dropout.

\paragraph{Data Sampling Strategy} For training all our models, we define an epoch as $X$ steps across the dataset, with $X=\frac{|\text{speech}\_\text{examples}|}{\text{batch}\_\text{size}}$. 
To construct the training data for each epoch, we sample batches by first applying the predefined task-level sampling ratios, followed by sampling based on the internal domain-level splits within each task.
In the case of multimodal training (speech and text tasks mixed), we consider speech as our \textit{main} modality, using it for defining epoch size and task ratio. Each time we sample a task and language split that has a textual equivalent (e.g., ST corresponds to MT; SQA to QA), we also sample a batch from the corresponding textual task. In practice, this means that every time a batch from ST en-de split is sampled, a batch from MT en-de follows it. We hypothesize that interleaving similar speech and textual batches during training provides regularization benefits, with the text data serving as a stabilizing signal for learning~\cite{pikabea2025breaking}.

\section{Experimental Settings}\label{sec:settings}

\paragraph{Codebase} We train our models using an internal fork of \texttt{torchtune}~\cite{torchtune}, which allows us to process interleaved representations of text and high-dimensional vectors within the user turn during instruction tuning.
The high-dimensional vectors pass through our speech projector, while the text prefix and suffix user prompts are processed by the LLM embedding layer. The obtained speech and text embeddings are both concatenated and fed into the first layer of the LLM which is trained on the masked input with standard cross-entropy loss. 
Different learning rate schedulers and optimizers are employed for the speech projector and the LoRA weights, allowing for more controlled and effective training of these distinct model components.

\paragraph{Inference Settings} 
We perform inference using \texttt{torchtune}, with a batch size of 1 and greedy decoding. The maximum number of new tokens was limited to 300. Unless explicitly stated otherwise, this decoding strategy was consistently applied across all experimental settings. Additional discussion regarding multimodal inference is present in Appendix Section~\ref{appendix:inferenceissues}.

\paragraph{Evaluation Metrics}
We evaluate our models on speech~(ASR, ST, SQA) and text~(MT, QA) tasks when relevant. For ASR, we score word error rate (WER) using HuggingFace evaluate library with default settings and MMS normalization~\cite{pratap2024scaling}. For ST/MT we present two evaluation metrics: BLEU4 computed with sacrebleu library~\cite{post-2018-call}\footnote{The signature is \small{``nrefs:1|case:mixed|eff:no|tok:\textbf{TOK}| smooth:exp|version:2.3.1''}, with $TOK=zh$ for Chinese, and $13a$ for the other languages.}, and COMET \cite{rei-etal-2022-comet}.\footnote{\texttt{Unbabel/wmt22-comet-da}} For SQA/QA, we use LLM-as-a-judge evaluation scripts  from the \texttt{bergen} library\footnote{\url{https://github.com/naver/bergen}}~\cite{rau-etal-2024-bergen}. We use their ``yes/no'' quality assessment evaluation format including the English reference text, the multilingual questions and the generated answers. We report average accuracy across four LLMs: \texttt{EuroLLM-9B-Instruct}~\cite{eurollm}, \texttt{Gemma3-12B/27B-Instruct}~\cite{gemma3}, and \texttt{Llama3.1-70B-Instruct}. 

\paragraph{Baselines} We compare our results with both backbones we use for training. We evaluate MT and QA using the reference transcripts and \llama{} in zero-shot settings, and we evaluate \seamless{} in both ASR and ST.

%% file: paper/4_results_section.tex
\section{Experiments}\label{sec:experiments}

We now present our results for ASR, ST, and SQA. Section~\ref{sec:experiments:models} introduces the models used in our experiments, followed by results and discussion in Section~\ref{sec:experiments:results}.

\subsection{Our Models}\label{sec:experiments:models}

In this section, we describe the models used in our experiments. Additional hyperparameter details are provided in Appendix~\ref{sec:appendix:hyperparameters}.

\paragraph{A.1 Speech Projector (ASR/ST)} 
This version of our speech projector focuses on ASR and ST tasks, which do not require any complex reasoning capability of the LLM. For ASR, we sample CoVoST2 and EuroParlST with probabilities 0.8 and 0.2 respectively. For ST, we sample proportionally to the datasets size, and set language sampling probabilities for the pairs en-de, en-zh, and en-it to 0.3, 0.4, and 0.3 respectively.  We trained for 4 epochs using AdamW with learning rate of $1e-4$, a constant learning rate scheduler, and gradient accumulation of 16. This model trains for 4.6 days in a single A100-80GB, and the best checkpoint is obtained after 18.2 hours.

\paragraph{A.2 Speech Projector (ASR/ST/SQA)} 
This version of our projector extends A.1, including the SQA task. We use task sampling probabilities of 0.4, 0.35, and 0.25 for ASR, ST and SQA respectively.  For the ASR and ST tasks, we use the same data ratios as defined above.
For the SQA task, we leveraged both valid and invalid splits, with language-specific sampling probabilities for English, German, and Italian set to 0.4, 0.3 and 0.3, respectively. We do not train with Chinese SQA. This model trains for 4.75 days in a single A100-80GB, and the best checkpoint is obtained after 27.36 hours. The best checkpoints for both versions of the speech projector are selected using its average ST performance on the ACL 60-60 dev split across all language directions. Additional hyper-parameters for A.1 and A.2 are presented in Appendix Section~\ref{appendix:modela:hyperparameters}.

\paragraph{B. Text-only LoRA (MT/QA)} We train LoRA weights on top of \llama{} using all text-to-text data from Table~\ref{tab:trainingdata}, and by using probability sampling of 0.6 and 0.4 for MT and QA respectively. We train for one epoch using AdamW with learning rate of $3e-4$, weight decay of $0.1$, and $100$ warm-up steps. Batch size of $10$, and gradient accumulation of $8$ is used. This model trains for approximately 4 days in a single A100-80GB. We select the last checkpoint.

\paragraph{C. Multimodal (A.x + B)} We restart training by using both one of the speech projectors detailed above, and the text-only LoRA weights. We adapt our models using all speech~(ASR/ST/SQA) and textual~(MT/QA) tasks. We experiment with two versions of the SQA/QA training sets: the original short lowercase answers, and the \textit{fluent} SQA/QA version we created.
In preliminary experiments, we observed that as little as 100 steps were enough to successfully integrate the projector representation to the LoRA weights, but the best performance gains were obtained with 1K steps, which is the value we adopt for the experiments presented in the next section. We use learning rate of $1e-5$ for the speech projector, and of $3e-4$ for the LoRA weights. We use a batch size of 16, and gradient accumulation of 16. This model trains for approximately 6 hours in a single A100-80GB. We select the last checkpoint.

\subsection{Results and Discussion}\label{sec:experiments:results}

\input{tables/results_main_table}

Table~\ref{tab:results} presents our results for ASR, ST/MT and SQA/QA. ACL 60-60 \textit{eval set} is used for ASR and ST/MT. SpokenSQuAD official test set is used for English SQA/QA. A smaller automatically obtained version is used for multilingual SQA/QA.\footnote{Statistics for the multilingual test set are presented in Appendix Table~\ref{tab:appspokensquadtestset}.} In the top portion, we present results for \llama{} before and after LoRA fine-tuning on text-only data. The middle portion of the table presents the speech backbone~(\seamless{}) and our projector-only models: for these rows, the only adaptation is the training of a speech projector that is plugged to a frozen \llama{}. Finally, the bottom portion of the table presents results from the merging of our text backbone (text-only LoRA) and the projectors of the middle portion via multimodal training. Additionally, ACL 60-60 ASR/ST dev results are presented in Appendix Table~\ref{tab:results-dev}.

\paragraph{Performance of Text-only Models (topline)} We observe that zero-shot \llama{} presents strong performance in both MT and QA tasks. By adding LoRA adapters on top of it, we increase translation performance in detriment of QA performance. We partially attribute this drop in QA performance to the SpokenSQuAD answer format, that is very short and might be judged as incomplete by the LLM evaluation. However, we also scored ROUGE1~\cite{lin-2004-rouge} recall, which measures the intersection between the reference answer tokens and the produced one, finding that those scores were similar to the LLM-as-a-judge metric.\footnote{For en, de, it and zh splits ROUGE1 recall scores were respectively: 81.4\%, 63.1\%, 69.5\%, 79.0\%.} This result confirms that the QA performance is worse after text adaptation.

\paragraph{Speech Projectors} We observe that both A.1 and A.2 are equally capable of performing ASR and ST, which is consistent with the fact that both are trained on the same data. However, we observe that A.2, the model that trains with SQA data, is unable to produce SQA output. We believe this is a limitation of the projector approach: we train a model capable of biasing the output of the LLM, which works well for content tasks such as ASR and ST. For a reasoning task, further adaptation might be required in order to force the model to comply to the instruction. Additional results for CoVoST2 and EuroParlST are presented in Appendix Table~\ref{tab:speech-projector}, and they confirm that both models are very similar in ST performance.

\paragraph{Multimodal Training} We observe that our efficient multimodal training is beneficial, consistently increasing scores for ASR and ST. 
Our multimodal models outperform the speech projector models by 1-2 WER points in the ASR task. For ST task, in BLEU scores, the multimodal models always outperform speech projector models while in COMET there are some mixed results, with some models presenting slight deterioration for some language pairs. Finally, in the SQA task, while the speech projector model A.2 failed to learn the task effectively, our multimodal models achieved strong results, outperforming the text topline (B) performance across all language settings, and reaching scores that are close to the \llama{} topline, despite working from the speech signal.

\paragraph{ASR Performance} We observe that our models are competitive with \seamless{}, scoring slightly worse than the baseline for some configurations.
Overall, for ACL 60-60 we observe quite elevated WER scores, compared to the ones we obtained for the training ASR datasets~(see Appendix Table~\ref{tab:speech-projector}). We believe this is partially due to the nature of the dataset. We manually inspected some of the data, observing that some of the audio files contain \textit{leftover fragments} from previous sentences. Moreover, looking at the transcriptions, we observed that these faithfully reproduced the audio, without applying any normalization or removing disfluencies~(e.g. repeated and filler words were kept). We find that disfluent transcriptions are hardly produced by LLM-based models or \seamless{}, that both tend to translate the content into a clean format. Therefore, we believe the WER scores presented in our results table are not representative of the models real ASR capabilities.

\paragraph{German MT/ST Performance} 
Across all experiments, we observed that our models consistently performed poorly on German, the language for which we had the largest amount of training data. This could be attributed to the LLM's inherent limitations in handling German, as reflected by its relatively lower performance in zero-shot settings~(\llama{}) for the en-de pair for both BLEU and COMET, compared to en-it and en-zh. Nonetheless, training the model on multilingual speech-only data for the ST task led to an improvement of 3–4 BLEU points. Incorporating multimodal data~(i.e., both ST and MT) yielded an additional gain of 2–3 BLEU points, further enhancing performance. For COMET, speech-only and multimodal training does not improve COMET scores over the text topline.

\paragraph{Overall ST Performance}
We observe that the speech projectors~(A.1 and A.2) outperform the backbone \seamless{} for en-zh using BLEU, and in all languages using COMET. Then, multimodal training further increases the BLEU scores, but in some cases, this training slightly hurts the COMET scores. Since this difference in COMET is very small~(en-de 0.005; en-it 0.008; en-zh 0.004), and since the BLEU scores increase, we attribute this to some formatting bias that could be happening during adaptation. The Appendix Section~\ref{appendix:modelc:hyperparameters} discusses the matter further.

\paragraph{SQA performance} Overall, we observe that by replacing SQA by \textit{fluent} SQA, we drastically increased our SQA/QA scores. However, results for ASR slightly deteriorate. We hypothesize that this is due to the old SQA task being closer to the ASR task. For the original SQA, the answer is always a direct transcript of a portion of the input text, which as a task has a better synergy with ASR. In intermediate experiments, we observed that adding the original SQA data to the multimodal training was always beneficial for the ASR performance of the model.

\paragraph{Final Discussion} Overall our results show that it is possible to train text~(B) and speech adaptation~(A) in parallel, and then to align both via joint instruction tuning~(C). By separating the pretraining of both components, we are able to focus hyper-parameter search at the merging stage, using two components that are already competent in their respective modalities. Despite improvements in scores over speech-only models, our best models do not beat the topline working from text for ST, but they do outperform \seamless{} across all language pairs and metrics. Moreover, for SQA, we highlight that the obtained scores are in some cases very close~(en-en, en-it) to the text topline, despite using speech as input context.

%% file: tables/results_main_table.tex
\begin{table*}
\centering
\resizebox{\textwidth}{!}{
\begin{tabular}{lc|ccc|ccc|cccc}\toprule
\multicolumn{1}{c}{}                   
& \textbf{ASR (WER)} 
& \multicolumn{3}{c|}{\textbf{ST/MT (BLEU)}}
& \multicolumn{3}{c|}{\textbf{ST/MT (COMET)}}
& \multicolumn{4}{c}{\textbf{SQA/QA (LLM-AS-A-JUDGE)}}                \\
\multicolumn{1}{l}{\textbf{Model (fine-tuning tasks)}}                   & \textbf{en}
& \textbf{en-de}
& \textbf{en-it}
& \textbf{en-zh}
& \textbf{en-de}
& \textbf{en-it}
& \textbf{en-zh}
& \textbf{en-en}
& \textbf{en-de}
& \textbf{en-it}
& \textbf{en-zh} \\\midrule
\multicolumn{12}{c}{\textbf{Text-only Models (MT/QA)}}\\\midrule
\llama{}~(zero-shot)   
& -
& 23.88
& 35.51
& 45.89
& 0.779
& 0.806
& 0.809
& \textbf{91.8\%}
& \textbf{92.0\%}
& \textbf{88.6\%}
& \textbf{84.6\%}         \\
\textbf{B.} Text-only LoRA~(MT/QA)
& -
& \textbf{41.69}
& \textbf{48.31}
& \textbf{53.65}
& \textbf{0.838}
& \textbf{0.863}
& \textbf{0.867}
& 83.4\%
& 75.7\%
& 71.4\%
& 69.5\%         \\\midrule
\multicolumn{12}{c}{\textbf{Speech-only Models (ASR/ST/SQA)}}\\\midrule
\seamless{} 
& \textbf{17.6}
& \textbf{27.95}
& \textbf{43.54}
& 33.58
& 0.737
& 0.788
& 0.753
& -
& -
& -
& -\\
\textbf{A.1} Speech Projector (ASR/ST)                           
& 19.8
& 27.58
& 36.30
& 40.62
& \textbf{0.760}
& 0.796
& \textbf{0.793}
& -
& -
& -
& -\\
\textbf{A.2} Speech Projector (ASR/ST/SQA)                       
& 19.9
& 27.20
& 36.60
& \textbf{40.72}
& \textbf{0.760}
& \textbf{0.797}
& 0.792
& 0.7\%
& 0.5\%
& 0.3\%
& 0.6\%          \\\midrule
\multicolumn{12}{c}{\textbf{Multimodal Models (ASR/ST/SQA)}}\\\midrule
\textbf{A.1 + B} (ASR/ST/MT/SQA/QA)
& \textbf{17.7}
& 30.37
& \textbf{41.22}
& 42.76
& 0.758
& \textbf{0.791}
& 0.795
& 79.8\%
& 71.9\%
& 69.4\%
& 65.5\%         \\
\textbf{A.1 + B} (ASR/ST/MT/\textit{fluent}SQA/\textit{fluent}QA)
& 18.6
& \textbf{30.75}
& 40.48
& 42.51
& 0.755
& 0.788
& 0.789
& 90.3\%
& 85.2\%
& 82.9\%
& 76.4\%         \\
\textbf{A.2 + B }(ASR/ST/MT/SQA/QA)
& 18.2
& 29.91
& 38.13
& 43.12
& 0.759
& 0.786
& \textbf{0.799}
& 80.5\%
& 74.9\%
& 68.0\%
& 66.7\%         \\
\textbf{A.2 + B }(ASR/ST/MT/\textit{fluent}SQA/\textit{fluent}QA) 
& 18.7
& 29.68
& 32.28
& \textbf{43.38}
& \textbf{0.763}
& 0.782
& 0.798
& \textbf{91.1\%}
& \textbf{87.3\%}
& \textbf{84.8\%}
& \textbf{78.0\%}        \\\bottomrule
\end{tabular}}
\caption{Results for the different models and backbones used in this work. ASR and ST scores are obtained using ACL 60-60 eval set, while SQA/QA scores are obtained using SpokenSQuAD test set.}
\label{tab:results}
\end{table*}

%% file: paper/5_submission_section.tex
\section{Submitted Model}\label{sec:submission}

Table~\ref{tab:results} presented the results for our multimodal models. Due to the reasons highlighted in Section~\ref{sec:datapreproc}, we only consider two models for submission: the ones which were trained with \textit{fluent} SQA. This is because we believe that these models will suffer the least from domain shift, since they are capable of producing full fluent sentences for SQA.

We observe that both models~(A.1 and A.2-based) seem to be equally capable of ASR~(18.6 and 18.7). We evaluated language confusion for these two splits, finding that the output produced was English in 98\% and 98.5\% of the cases respectively. These models differ more in terms of ST performance: they obtained averages of 37.9 and 35.1 BLEU score points respectively. Finally, these models present the following average accuracy for SQA: 83.7\% and 85.3\%. In summary, while the A.1-based model seem much more capable in ST, the A.2-based model has a very slight edge in ASR and SQA.

Therefore, we decided to select the A.1-based model as our primary submission model. For producing the decoding of the test set, we transform the instructions into our own prompt format, and submit the output of the same model, with the same decoding settings for all splits: greedy decoding using 150 as maximum number of tokens. A brief post-submission discussion is provided in Appendix Section~\ref{sec:appendix:postusbmission}.



%% file: paper/6_conclusion.tex
\section{Conclusion}\label{sec:conclusion}

In this paper, we presented NLE's submission to the instruction-following speech processing short track at IWSLT 2025, constrained setting. We developed multimodal models that simultaneously performed ASR, ST, and SQA tasks from English speech input into Chinese, German and Italian. Our approach is simple yet effective: we decoupled training, training a speech projector on speech-to-text tasks, and LoRA adapters on text-to-text tasks. Then, both modules are loaded and the resulting multimodal model was instruction-tuned for a few steps on multilingual and multimodal data to produce the final system submitted for evaluation.



%% file: supplementary/appendix_section.tex
\section{Data Preprocessing and Prompts}
\label{sec:appendix}

\input{tables/regeneration_prompt}

\input{tables/prompt_format}

\subsection{SpokenSQuAD TTS data}\label{app:spokensquadtts}

In this section we explain the misalignment we found in the SpokenSQuAD train split, as well as our procedure for creating the new TTS split.

\paragraph{Train Split Misalignment} During the preprocessing of the train split of SpokenSQuAD, we witnessed cases of misalignment: for some given paragraphs, the corresponding audios were shifted by a factor varying between 1 and 3. For instance, the first audio in a given paragraph was incorrectly named as ``1'', instead of ``0'', shifting all the paragraph's alignment. We listened to all cases we were able to flag, manually correcting them. However, we believe this hinted to a deeper alignment issue, as the obtained training set seemed to be difficult to learn. We observed that models trained with this training set included were unable to generalize to SpokenSQuAD's validation and test sets, always producing random Wikipedia sentences when receiving the SpokenSQuAD's TTS voice as input.

\paragraph{New TTS Source Audio Generation} We use \seamless{} to produce new source audio for the SpokenSQuAD training set~(34,003 sentences). For each entry in this set, we re-synthesize its SQuAD reference text by randomly sampling one of the 200 speakers present in \texttt{SeamlessM4T}. This results in a more diverse training set, since the original TTS used a single female voice for all sentences. We also generated extra speech data using all different questions present in this training set, producing a second ASR set containing speech for 28,000 questions.

\subsection{SpokenSQuAD Answers Regeneration}\label{app:spokensquadnewanswers}

SpokenSQuAD answers are direct extracts from the reference text, formatted as lowercase text without punctuation. We discovered this presented a limitation when training our models on SQA. The trained models over-fitted to that format, solving SQA as a transcription task of the relevant portion of the source audio. While conceptually correct, this approach can result in generalization issues if more than one extract is required to answer the question, as the model never observed such a setting during training, and it will thus have the tendency to transcribe everything between the two points of interest.

We use \llama{} to regenerate all answers in our multilingual training set. The prompt used for regeneration is presented in Table~\ref{tab:regenerationprompt}. After regeneration, we remove answers generated in the wrong language using an automatic language identification tool. Statistics are presented in Table~\ref{tab:trainingdata}.

\subsection{Data Statistics}

Table~\ref{tab:appspokensquadtestset} presents the statistics for the multilingual test set of SpokenSQuAD.

\subsection{Our Prompts}

Table~\ref{tab:prompts} presents our prompt format. We designed our prompts to be very similar, independently of the target task. The language of the question defines the answer's target language.


\section{Additional Results}

\paragraph{CoVoST2 and EuroParlST Results} Table~\ref{tab:speech-projector} presents results for relevant models on the in-domain test sets from CoVoST2 and EuroParlST. We observe that the multimodal adaptation improves the speech projectors' ASR and BLEU scores, while slightly decreasing COMET scores.

\paragraph{ACL 60-60 Dev Results} Table~\ref{tab:results-dev} presents ACL~60-60 dev split scores for some of the models presented in the main results table~(Table~\ref{tab:results}).

\paragraph{SQA/QA BERT Scores} Table~\ref{appendix:bertscore:valid} presents BERT scores for the multimodal models presented in the main results table~(Table~\ref{tab:results}), computed after the evaluation period and using the same settings from \citet{abdulmumin-etal-2025-findings}. We observe that scores for languages other than English over the valid test set are considerably lower than the LLM-as-judge scores in Table~\ref{tab:results}. Prior to the evaluation period we had evaluated our models using BERT score and \texttt{xlm-roberta-large}, which yielded much higher scores, similar to those obtained in the LLM-as-judge evaluation. Those scores are presented in Table~\ref{appendix:bertscore:valid2}.

\paragraph{Invalid Questions} In Table~\ref{appendix:bertscore:valid} we also present BERT scores for an invalid multilingual test set, made of the same reference audio files from the English test split~(8,026 examples), but with incorrect questions. We observe that the BERT scores for this invalid split is very high, showcasing that our models are fully capable of respecting the instruction format for incorrect answers.

\input{tables/spokensquad_testset}

\subsection{Inference Issues}\label{appendix:inferenceissues}

Through manual inspection of our model outputs, we observed that in a small number of cases, inference degenerates, resulting in repeated words or sentences until the maximum token limit is reached. We experimented with various inference strategies, greedy decoding, top-p, and top-k sampling, as well as different temperature settings, but were unable to identify a configuration that fully eliminated the issue. We hypothesize that a lightweight post-processing model could offer a simple and effective solution to mitigate this problem. Below we give some examples of inference degeneration for Chinese and German.

An example of inference degeneration in Chinese:\\
\indent``\begin{CJK*}{UTF8}{gbsn}国家语言理解模型从各种知识来源中提取出来，例如，\fix{通常通过预训练获得的参数中包含的知识，通常通过预训练获得的知识，通常通过预训练获得的知识，通常通过预训练获得的知识，}(..)\fix{通常通过预训练}\end{CJK*}''

An example of inference degeneration in German:\\
\indent\textit{``In Zusammenhang mit der semantischen Parsierung, wenn wir nach der kompositionalen Generalisierung testen, könnte es mir so vorkommen, als ob wir in diesem Fall die Mädchen schlafen, \fix{und ich sehe, dass sie schlafen, und ich sehe, dass sie schlafen,} (..) \fix{und ich sehe, dass sie schlafen, und ich sehe, dass}''}

\input{tables/speech_projector_result}
\input{tables/results_dev}
\input{tables/appendix_bert_scores_valid}
\input{tables/appendix_bert_scores_2}

\clearpage
\section{Models Hyperparameters}\label{sec:appendix:hyperparameters}

\input{tables/appendix_multimodal_ablation}
\input{tables/appendix_multimodal_ablation2}

Table~\ref{tab:listdatasets} lists the data splits used for each model presented in results Table~\ref{tab:results}. Table~\ref{tab:probabilityinfo} presents the probability sampling employed during training.

\input{tables/data_per_setup}
\input{tables/probability_parameters}

\subsection{Speech Projector (A) Hyperparameters}\label{appendix:modela:hyperparameters}

\paragraph{Architecture} 
We explored multiple architectures to map speech embeddings from \seamless{} to \llama{}. To train the speech projector, speech features extracted from \seamless{} are input to the projector, and the resulting outputs are passed through a frozen \llama{} model. The speech projector, initialized with random weights, is trained using cross-entropy loss on an ASR task.
Preliminary experiments demonstrated that the Transformer encoder architecture consistently outperformed both the Conformer and Multi-Layer Perceptron architectures of similar parameter sizes. 
Consequently, we adopt the Transformer encoder architecture for all experiments presented in this work.

\paragraph{Averaged Features} As mentioned, in preliminary experiments, we experimented using the original output of \seamless{}, as well as performing average every 2 or 3 frames. We observe that averaging every 3 frames results in models that are considerably faster to train, while maintaining similar performance to the original output.

\paragraph{Data Ratio} For the ASR task, preliminary experiments revealed that training solely on EuroparlST ASR data resulted in poor generalization, whereas incorporating CoVoST ASR data significantly improved model robustness. 
For the ST task, we defined the data sampling ratios according to the target language distribution across the CoVoST and EuroparlST datasets.

\paragraph{Batch Size} ASR and ST tasks use a batch size of 16, while SQA is batched with size 8, due to the longer user prompts.

\paragraph{Checkpoint Selection} Checkpoints were selected based on development set performance across three or four configurations: ASR-best, ST-best, SQA-best~(A.2-only), and an All-best checkpoint combining all tasks. We only present results for ST-best checkpoints, which we found to produce the best scores in ST compared to the other versions, while only marginally decreasing scores in ASR compared to the ASR-best checkpoint. We do not consider SQA-best checkpoints, as the overall SQA performance of projector-only models is very low regardless of the checkpoint selection method.

\paragraph{Exclusion of Chinese SQA Data} During the training of the speech projector~(A.2), we excluded Chinese SQA data. This was due to parallel observation in B models~(text-only), in which we observed that the LLM failed to generate coherent Chinese answers. While later we were able to confirm that the issue did not come from the Chinese split itself, this model was obtained simultaneously to that finding, explaining why the data was not included in this setting.

\subsection{Multimodal Models (C) Hyperparameters}\label{appendix:modelc:hyperparameters}

In this section we present some ablation experiments for our multimodal adaptation setup. The experiments are performed by producing variants of the \textit{A.1+B} model, which is the model we submitted to the challenge. Table~\ref{tab:appendix:multimodalablation} present ACL~60-60 dev split ASR and ST results that are discussed in the next paragraphs.

\paragraph{Impact of Parameters Count} During our multimodal merging step, we combine text-only LoRA weights with our speech projector, yielding better scores. Since this increase in scores could be simply due the additional parameters, we trained a variant of our model in which the merging step is performed using a randomly initialized LoRA. We observe that our training setup indeed benefits from \textit{any} additional weights during adaptation: the models trained with a randomly initialized LoRA outperform the speech projector backbone~(A.1). Adding textual tasks in this setting does not help the system, which we hypothesize is due to the LoRA weights not being pretrained on the textual task.
Finally, adapting using a pretrained LoRA model further improves ASR and ST scores for two out of three language directions~(en-it and en-zh).

\paragraph{Impact of Textual Tasks} For the multimodal models presented in the main paper, we adapt pretrained modules leveraging speech and textual tasks. We thus investigated the impact of having aligned speech and textual tasks during this adaptation.
We observe that incorporating textual tasks has little impact on ASR performance, while substantially improving ST performance for Italian. The results are less favorable for German and Chinese: in German, the addition of textual tasks leads to a performance drop, whereas in Chinese, the decline is minimal. Overall, these findings suggest that the textual modality may be particularly beneficial in low-resource settings. Italian, which has the fewest training examples in our dataset, appears to benefit the most from this adaptation.

\paragraph{Impact of SQA} Examining the results in Table~\ref{tab:appendix:multimodalablation}, we observe that incorporating the SQA and QA tasks leads to improvements in both ASR and ST performance. We hypothesize that the SQA task enhances the model's adherence to the prompt by encouraging it to \textit{attend to} the information provided, thereby reducing both task and language confusion.

\paragraph{Inclusion of Synthetic Textual Data} 
Table~\ref{tab:appendix:multimodalablation2} presents the results of our investigation into the inclusion of potentially noisy synthetic textual data during training. We observe that excluding this synthetic data (\textit{No synthetic data}) negatively affects both BLEU and WER scores. Conversely, training exclusively with synthetic data (\textit{Only synthetic data}) yields improved performance across all metrics. We attribute this to the fact that the target text in the non-synthetic data is duplicated from the speech task~(i.e. the MT set is built from the ST set), leading to reduced data diversity. Removing this duplicated data introduces greater variability during the adaptation phase, which appears beneficial. Finally, using both types of data leads to improved performance for all languages except German. As discussed in the main paper, we hypothesize that this discrepancy is due to issues in the German training data sourced from EuroParlST and CoVoST2.

\paragraph{Number of Adaptation Steps} Table~\ref{tab:appendix:multimodalablation} presents results for a version of our model trained for twice as long (2K steps). At this point, we observe signs of training saturation: differences in ASR and COMET scores across all metrics are minimal, and BLEU scores drop for both German and Italian. These results suggest that our adaptation step does not require a considerable number of training steps. 

\paragraph{Task Ratios} On our preliminary experiments, we tested different task ratios, selecting the one with best average WER and BLEU scores over both ACL~60-60 dev and test set. Table~\ref{tab:appendix:multimodalablation2} presents those results.

\subsubsection{Post-submission Discussion} \label{sec:appendix:postusbmission}
Due to time constraints, many of our ablation studies were conducted after the initial submission. Upon analyzing these results~(Table~\ref{tab:appendix:multimodalablation}), we hypothesize that there may be an issue with the German training data: the more the model is exposed to it during training, the worse the COMET scores become. This hypothesis is supported by our \textit{Only synthetic data} results, which show improved BLEU scores for German when we exclude textual data from EuroParlST and CoVoST2. Additionally, our ablations suggest that using textual data selectively, rather than uniformly, may be more effective. In particular, textual supervision appears to be most beneficial for Italian, with more limited gains observed for the other two language directions. 

Regarding SQA, we were surprised to find that the evaluation setup provided by the organizers yields scores that differ significantly from those obtained with our own evaluation protocol (see Tables~\ref{appendix:bertscore:valid} and \ref{appendix:bertscore:valid2}). These discrepancies also extend to the scores we obtain using LLM-as-judge~(Table~\ref{tab:results}). We plan to further investigate the limitations of our current evaluation setup to better understand these inconsistencies.

%% file: tables/regeneration_prompt.tex
\begin{table*}
\resizebox{\textwidth}{!}{
\begin{tabular}{l}\toprule
Context: \textbf{{[}REFERENCE TEXT{]}} \\
Question: \textbf{{[}QUESTION{]}}\\
Answer: \textbf{{[}ANSWER{]}}\\
\begin{tabular}[c]{@{}l@{}}Instruction: Reformulate the answer to be a natural sounding sentence that answers the question in the correct language. \\ Produce text in the same language of the question and answer. Do not make it too long, or add too much information. \\ Don't add anything else to your answer.\end{tabular}\\\bottomrule
\end{tabular}}
\caption{Regeneration prompt we gave to \llama{} to regenerate the answers in the training set of SpokenSQuAD.}
\label{tab:regenerationprompt}
\end{table*}

%% file: tables/prompt_format.tex
\begin{table*}[]
\centering
\begin{tabular}{cl}\toprule
\textbf{}        & \multicolumn{1}{c}{\textbf{User Prompt}}\\\midrule
\multicolumn{1}{l}{} \textbf{Speech Prefix} & Content: \textless{}speech\textgreater{}{[}\textbf{SPEECH EMBEDDINGS}{]}\textless{}/speech\textgreater{}\textbackslash{}n\\\midrule
\multicolumn{1}{l}{} \textbf{Text Prefix} & Content: \textless{}text\textgreater{}{[}\textbf{SPEECH TRANSCRIPTION}{]}\textless{}/text\textgreater{}\textbackslash{}n\\\midrule
\textbf{ASR}& \begin{tabular}[c]{@{}l@{}}Question: Can you transcribe the Speech content into English text?\textbackslash{}n\end{tabular}\\\midrule
\textbf{ST/MT (de)}& \begin{tabular}[c]{@{}l@{}}Question: Können Sie den Inhalt der Rede in den deutschen Text übersetzen?\textbackslash{}n\end{tabular} \\
\textbf{ST/MT (it)}& \begin{tabular}[c]{@{}l@{}}Question: Puoi tradurre il contenuto del discorso in testo italiano?\textbackslash{}n\end{tabular} \\
\textbf{ST/MT (zh)}& \begin{tabular}[c]{@{}l@{}}Question: \begin{CJK*}{UTF8}{gbsn}你能把演讲内容翻译成中文吗?\textbackslash{}n\end{CJK*}\end{tabular} \\
\midrule
\textbf{SQA/QA}& \begin{tabular}[c]{@{}l@{}}Question: {[}\textbf{QUESTION}{]}\textbackslash{}n\end{tabular}\\\midrule
\textbf{Suffix} & Your answer: 
\\\bottomrule
\end{tabular}
\caption{The user turn prompt template used for training our models. For speech tasks, the user prompt is given by \textbf{Speech Prefix+Task+Suffix}, for text tasks, the user prompt is given by \textbf{Text Prefix+Task+Suffix}. \\ST/MT instructions were obtained by translating the instruction ``\textit{Can you translate the Speech content into \textbf{{[}German/Italian/Chinese{]}} text?}'' to corresponding target languages using \seamless{}.}
\label{tab:prompts}
\end{table*}

%% file: tables/spokensquad_testset.tex
\begin{table}
\centering
\begin{tabular}{ccccc}\toprule
                  & \textbf{en} & \textbf{de} & \textbf{it} & \textbf{zh} \\\midrule
\textbf{\# lines} & 8,026       & 3,272       & 783         & 627    \\\bottomrule
\end{tabular}
\caption{Statistics for the SpokenSQuAD SQA/QA test sets. The multilingual version is obtained via \seamless{} translation of questions and answers, with posterior quality filtering based on COMET scores.}
\label{tab:appspokensquadtestset}
\end{table}

%% file: tables/speech_projector_result.tex
\begin{table*}
\centering
\resizebox{\textwidth}{!}{%
\begin{tabular}{l|cc|cccccccccc}\toprule
\multicolumn{1}{l}{} &
  \multicolumn{4}{|c|}{\textbf{ASR}} &
  \multicolumn{8}{c}{\textbf{ST}} \\ \midrule 
\multicolumn{1}{l}{} &
  \multicolumn{2}{|c}{\textbf{CoVoST2}} &
  \multicolumn{2}{|c|}{\textbf{EuroParlST}} &
  \multicolumn{4}{c|}{\textbf{CoVoST2}} &
  \multicolumn{4}{c}{\textbf{EuroParlST}} \\ 
\multicolumn{1}{l}{} &
  \multicolumn{2}{|c}{} & \multicolumn{2}{|c|}{}
  &
  \multicolumn{2}{c|}{\textbf{en-de}} &
  \multicolumn{2}{c|}{\textbf{en-zh}} &
  \multicolumn{2}{c|}{\textbf{en-de}} &
  \multicolumn{2}{c}{\textbf{en-it}} \\
\multicolumn{1}{l|}{} &
  WER &
  \multicolumn{1}{c|}{CER} &
  WER &
  \multicolumn{1}{c|}{CER} &
  BLEU &
  \multicolumn{1}{c|}{COMET} &
  BLEU &
  \multicolumn{1}{c|}{COMET} &
  BLEU &
  \multicolumn{1}{c|}{COMET} &
  BLEU &
  COMET \\ \midrule
\multicolumn{1}{l|}{\textbf{A.1} (ASR/ST)} &
  7.84 &
  \multicolumn{1}{c|}{3.70} &
  11.21 &
  \multicolumn{1}{c|}{7.30} &
  \textbf{30.24} &
  \multicolumn{1}{c|}{\textbf{0.789}} &
  \textbf{40.14} &
  \multicolumn{1}{c|}{\textbf{0.806}} &
  \textbf{25.06} &
  \multicolumn{1}{c|}{\textbf{0.840}} &
  \textbf{27.55} &
  0.860 \\ 
\multicolumn{1}{l|}{\textbf{A.2} (ASR/ST/SQA)} &
  \textbf{7.21} &
  \multicolumn{1}{c|}{\textbf{3.44}} &
  \textbf{10.98} &
  \multicolumn{1}{c|}{\textbf{7.18}} &
  30.68 &
  \multicolumn{1}{c|}{\textbf{0.789}} &
  40.63 &
  \multicolumn{1}{c|}{0.807} &
  25.21 &
  \multicolumn{1}{c|}{0.841} &
  28.36 &
  \textbf{0.859} \\\midrule
  \textbf{A.1 + B} (ASR/ST/MT/\textit{fluent}SQA/\textit{fluent}QA) & 7.59	& 3.55	&11.20	& 7.20 &	\multicolumn{1}{|c}{31.42} &	0.770	& \multicolumn{1}{|c}{42.19}	& 0.802	& \multicolumn{1}{|c}{27.16} &	0.809	& \multicolumn{1}{|c}{27.86}	& 0.842
  \\
  \textbf{A.2 + B} (ASR/ST/MT/\textit{fluent}SQA/\textit{fluent}QA) & 7.01 &	3.25 &	10.82	 & 7.11	& \multicolumn{1}{|c}{31.83} &	0.772	& \multicolumn{1}{|c}{42.36}	& 0.804	& \multicolumn{1}{|c}{26.70}	&0.812	&\multicolumn{1}{|c}{27.77}	&0.848 \\
  \bottomrule
\end{tabular}%
}
\caption{ASR and ST results for the test sets of CoVoST2 and EuroParlST.}
\label{tab:speech-projector}
\end{table*}

%% file: tables/results_dev.tex
\begin{table*}
\centering
\resizebox{\textwidth}{!}{
\begin{tabular}{lc|ccc|ccc}\toprule
\multicolumn{1}{c}{}& \textbf{ASR (WER)} & \multicolumn{3}{c|}{\textbf{ST/MT (BLEU)}}        & \multicolumn{3}{c}{\textbf{ST/MT (COMET)}}\\
\multicolumn{1}{l}{\textbf{Model (fine-tuning data)}} & \textbf{en}  & \textbf{en-de} & \textbf{en-it} & \textbf{en-zh} & \textbf{en-de} & \textbf{en-it} & \textbf{en-zh}\\\midrule
\multicolumn{8}{c}{\textbf{Text-only models (MT/QA)}}\\\midrule
\llama{}~(zero-shot)   
& -
& 21.27
& 33.81
& 44.01
& 0.732
& 0.757
& 0.755
\\
\textbf{B} Text-only LoRA~(MT/QA)
& -
& \textbf{36.94}
& \textbf{49.81}
& \textbf{52.33}
& \textbf{0.782}
& \textbf{0.815}
& \textbf{0.822}
\\\midrule
\multicolumn{8}{c}{\textbf{Speech-only models (ASR/ST/SQA)}}\\\midrule
\seamless{}
& 25.3
& 23.77
& \textbf{37.84}
& 28.17
& 0.669
& 0.713
& 0.663
\\
\textbf{A.1} Speech Projector (ASR/ST)                           
& \textbf{15.9}
& 26.77
& 36.34
& 38.65
& 0.718
& \textbf{0.750}
& \textbf{0.753}
\\
\textbf{A.2} Speech Projector (ASR/ST/SQA)                       
& \textbf{15.9}         
& \textbf{26.85}
& 36.09
& \textbf{38.85}
& \textbf{0.720}
& 0.749
& \textbf{0.753}
\\\midrule
\multicolumn{8}{c}{\textbf{Multimodal models (ASR/ST/SQA)}}\\\midrule
\textbf{A.1 + B} (ASR/ST/MT/SQA/QA)
& \textbf{13.9}
& 28.74
& 41.73
& 40.72
& 0.716
& \textbf{0.759}
& 0.756
\\
\textbf{A.1 + B} (ASR/ST/MT/\textit{fluent}SQA/\textit{fluent}QA) 
& 14.4
& 27.20
& \textbf{42.01}
& 41.23
& 0.712
& 0.752
& 0.753
\\
\textbf{A.2 + B }(ASR/ST/MT/SQA/QA)
& 17.2
& \textbf{29.18}
& 39.14
& 40.99
& 0.719
& 0.755
& 0.762
\\
\textbf{A.2 + B} (ASR/ST/MT/\textit{fluent}SQA/\textit{fluent}QA)
& 15.5         
& 27.62
& 33.22          
& \textbf{41.74}          
& \textbf{0.726}          
& 0.749          
& \textbf{0.764}          
\\\bottomrule
\end{tabular}}
\caption{ACL 60-60 dev set results for the different models and backbones used in this work.}
\label{tab:results-dev}
\end{table*}

%% file: tables/appendix_bert_scores_valid.tex
\begin{table*}
\centering
\resizebox{\textwidth}{!}{
\begin{tabular}{ccccc|cccc}\toprule
    & \multicolumn{4}{c}{\textbf{Valid Questions}}                      & \multicolumn{4}{c}{\textbf{Invalid Questions}}                    \\
    & \textbf{en-en} & \textbf{en-de} & \textbf{en-it} & \textbf{en-zh} & \textbf{en-en} & \textbf{en-de} & \textbf{en-it} & \textbf{en-zh} \\\midrule
\textbf{A.1 + B} (ASR/ST/MT/\textit{fluent}SQA/\textit{fluent}QA) & 0.975          & 0.532          & 0.541          & 0.665          & 0.999          & 0.984          & 0.990          & 0.988          \\
\textbf{A.2 + B} (ASR/ST/MT/\textit{fluent}SQA/\textit{fluent}QA) & 0.975          & 0.536          & 0.546          & 0.666          & 0.999          & 0.990          & 0.989          & 0.991         \\\bottomrule
\end{tabular}}
\caption{BERT scores for SpokenSQuAD test sets computed using the same settings from the organizers~(default model, bert\_score version~0.3.13). 
Invalid questions corresponds to a version of the test set in which the questions are impossible to answer given the speech context.}
\label{appendix:bertscore:valid}
\end{table*}

%% file: tables/appendix_bert_scores_2.tex
\begin{table*}
\centering
\resizebox{\textwidth}{!}{
\begin{tabular}{ccccc|cccc}\toprule
    & \multicolumn{4}{c}{\textbf{Valid Questions}}                      & \multicolumn{4}{c}{\textbf{Invalid Questions}}                    \\
    & \textbf{en-en} & \textbf{en-de} & \textbf{en-it} & \textbf{en-zh} & \textbf{en-en} & \textbf{en-de} & \textbf{en-it} & \textbf{en-zh} \\\midrule
\textbf{A.1 + B} (ASR/ST/MT/\textit{fluent}SQA/\textit{fluent}QA) & 0.857          & 0.869          & 0.863          & 0.899          & 0.996          & 0.997          & 0.996          & 0.997          \\
\textbf{A.2 + B} (ASR/ST/MT/\textit{fluent}SQA/\textit{fluent}QA) & 0.857          & 0.869          & 0.863          & 0.896          & 0.995          & 0.998          & 0.998          & 0.998   \\\bottomrule      
\end{tabular}}
\caption{BERT scores for SpokenSQuAD test sets computed using \texttt{xlm-roberta-large}. All other settings are equal to Table~\ref{appendix:bertscore:valid}. Invalid questions corresponds to a version of the test set in which the questions are impossible to answer given the speech context.}
\label{appendix:bertscore:valid2}
\end{table*}

%% file: tables/appendix_multimodal_ablation.tex
\begin{table*}
\centering
\resizebox{\textwidth}{!}{\begin{tabular}{lcc|ccc|ccc}\toprule
\multicolumn{1}{c}{}                     & \multicolumn{2}{c}{\textbf{ASR}} & \multicolumn{3}{c}{\textbf{ST (BLEU)}}           & \multicolumn{3}{c}{\textbf{ST (COMET)}}          \\
\multicolumn{1}{c}{}                     & \textbf{WER}    & \textbf{CER}   & \textbf{en-de} & \textbf{en-it} & \textbf{en-zh} & \textbf{en-de} & \textbf{en-it} & \textbf{en-zh} \\\midrule
\textbf{A.1} (ASR/ST)                         & 15.9            & 7.7            & 26.77          & 36.34          & 38.65          & 0.718          & 0.750          & 0.753          \\\midrule
\textbf{A.1 + random LoRA} (ASR/ST/SQA)           & 17.4            & 8.2            & \textbf{29.07} & 37.04          & \textbf{41.47}          & \textbf{0.732}          & \textbf{0.760} & \textbf{0.761}          \\
\textbf{A.1 + random LoRA} (ASR/ST/SQA) + (MT/QA) & \textbf{16.4}            & \textbf{7.7}            & 28.07          & \textbf{38.88}          & 41.24          & 0.725          & 0.755          & 0.760          \\\midrule
\textbf{A.1 + B} (ASR/ST) & \textbf{13.8}	& \textbf{6.4} & \textbf{27.91}	& 28.88 & 	\textbf{41.70}	& \textbf{0.728}	& \textbf{0.743}	& \textbf{0.760} \\

\textbf{A.1 + B} (ASR/ST) + (MT) &14.9	&7.2 &26.09	&\textbf{32.97}	&41.10	&0.715	&0.741	&0.755 \\\midrule
\textbf{A.1 + B} (ASR/ST/SQA) & \textbf{14.1}	&\textbf{6.5} &\textbf{29.02}	& 30.08	& \textbf{41.85}	&\textbf{0.722}	&0.743	&\textbf{0.763} \\
\textbf{A.1 + B} (ASR/ST/SQA) + (MT/QA)           & 14.4        & \textbf{6.5}         & 27.20       & \textbf{42.01}       & 41.23       & 0.712                & \textbf{0.752}                & 0.753                \\\midrule
\textbf{A.1 + B} (ASR/ST/SQA) + (MT/QA) \textbf{No synthetic data}  &18.8 & 9.2 & 28.47	& 32.97	& 40.03	& 0.717	& 0.741 &	0.753\\
\textbf{A.1 + B} (ASR/ST/SQA) + (MT/QA) \textbf{Only synthetic data}  & \textbf{13.9}	& \textbf{6.6} & \textbf{29.71}	& \textbf{39.80}	& \textbf{42.04}	& \textbf{0.721}	& \textbf{0.751}	& \textbf{0.754} \\\midrule
\textbf{A.1 + B} (ASR/ST/SQA) + (MT/QA) \textbf{2K steps}  & 14.3	& 6.4&  27.09	& 40.53 & 	41.53	& 0.713	& 0.753	& 0.755 \\\bottomrule
\end{tabular}}
\caption{ACL~60-60 dev set ASR and ST scores for variants of our best model (A.1+B).}
\label{tab:appendix:multimodalablation}
\end{table*}

%% file: tables/appendix_multimodal_ablation2.tex
\begin{table}
\centering
\resizebox{\columnwidth}{!}{
\begin{tabular}{ccc}\toprule
 ASR/ST/SQA                 & \textbf{average WER} & \textbf{average BLEU} \\\midrule
0.2 / 0.4 / 0.4   & \textbf{16.03}       & \textbf{37.58}        \\
0.2 / 0.5 / 0.3   & 17.72                & 36.36                 \\
0.2 / 0.6 / 0.2   & 16.24                & 36.62                 \\
0.25 / 0.5 / 0.25 & 16.58                & 37.08                 \\
0.3 / 0.5 / 0.2   & 16.23                & 37.30                \\\bottomrule
\end{tabular}}
\caption{ACL~60-60 dev and test set average WER and BLEU scores for our best model (A.1+B) by varying the ASR/ST/SQA ratios.}
\label{tab:appendix:multimodalablation2}
\end{table}

%% file: tables/data_per_setup.tex
\begin{table*}
\centering
\resizebox{\textwidth}{!}{
\begin{tabular}{lccc|ccc|cccc}\toprule
& \multicolumn{3}{c}{\textbf{CoVoST}}      & \multicolumn{3}{c}{\textbf{EuroParlST}}  & \multicolumn{4}{c}{\textbf{SpokenSQuAD}}                              \\
& \textbf{ASR} & \textbf{ST} & \textbf{MT} & \textbf{ASR} & \textbf{ST} & \textbf{MT} & \textbf{ASR} & \textbf{MT} & \textbf{SQA/QA} & \textbf{\textit{fluent} SQA/QA} \\\midrule
\textbf{B} Text-only LoRA
&  $$ \ding{55} $$  
&  $$ \ding{55} $$  
& $$ \ding{51} $$   
&   $$ \ding{55} $$    
&  $$ \ding{55} $$    
& $$ \ding{51} $$    
&    $$ \ding{55} $$     
& $$ \ding{51} $$   
& $$ \ding{51} $$  
&   $$ \ding{55} $$     \\
\textbf{A.1} Speech Projector (ASR/ST)     
& $$ \ding{51} $$   
& $$ \ding{51} $$  
&   $$ \ding{55} $$   
& $$ \ding{51} $$   
& $$ \ding{51} $$    
& $$ \ding{55} $$ 
& $$ \ding{51} $$ 
&  $$ \ding{55} $$ 
&  $$ \ding{55} $$  
&  $$ \ding{55} $$   \\
\textbf{A.2} Speech Projector (ASR/ST/SQA) 
& $$ \ding{51} $$    
& $$ \ding{51} $$    
&  $$ \ding{55} $$  
& $$ \ding{51} $$    
& $$ \ding{51} $$   
&  $$ \ding{55} $$  
&  $$ \ding{51} $$  
&  $$ \ding{55} $$ 
& $$ \ding{51} $$ (no zh)    
&  $$ \ding{55} $$   \\
\textbf{A.1 + B} Multimodal model (ASR/ST/MT/SQA/QA)
&  $$ \ding{51} $$  
&  $$ \ding{51} $$ 
&  $$ \ding{51} $$ 
&  $$ \ding{51} $$  
& $$ \ding{51} $$   
& $$ \ding{51} $$  
& $$ \ding{51} $$  
&  $$ \ding{51} $$ 
&  $$ \ding{51} $$   
&  $$ \ding{55} $$\\
\textbf{A.1 + B} Multimodal model (ASR/ST/MT/fluentSQA/fluentQA)
&  $$ \ding{51} $$  
&  $$ \ding{51} $$ 
&  $$ \ding{51} $$ 
&  $$ \ding{51} $$  
& $$ \ding{51} $$   
& $$ \ding{51} $$  
& $$ \ding{51} $$  
&  $$ \ding{51} $$ 
&  $$ \ding{55} $$   
&  $$ \ding{51} $$\\
\textbf{A.2 + B} Multimodal model (ASR/ST/MT/SQA/QA)
&  $$ \ding{51} $$  
&  $$ \ding{51} $$ 
&  $$ \ding{51} $$ 
&  $$ \ding{51} $$  
& $$ \ding{51} $$   
& $$ \ding{51} $$  
& $$ \ding{51} $$  
&  $$ \ding{51} $$ 
&  $$ \ding{51} $$   
&  $$ \ding{55} $$\\
\textbf{A.2 + B} Multimodal model (ASR/ST/MT/fluentSQA/fluentQA)
&  $$ \ding{51} $$  
&  $$ \ding{51} $$ 
&  $$ \ding{51} $$ 
&  $$ \ding{51} $$  
& $$ \ding{51} $$   
& $$ \ding{51} $$  
& $$ \ding{51} $$  
&  $$ \ding{51} $$ 
&  $$ \ding{55} $$   
&  $$ \ding{51} $$\\\bottomrule                  
\end{tabular}}
\caption{List of datasets and splits used for each model presented in Table~\ref{tab:results}. Statistics for number of examples can be seen in Table~\ref{tab:trainingdata}.}
\label{tab:listdatasets}
\end{table*}

%% file: tables/probability_parameters.tex
\begin{table*}
\centering
\resizebox{\textwidth}{!}{
\begin{tabular}{lc|cccc|ccccc}\toprule
& \textbf{ASR}
& \multicolumn{4}{c|}{\textbf{ST/MT}}
& \multicolumn{5}{c}{\textbf{SQA/QA (valid/invalid)}}                           
\\
& \textbf{\begin{tabular}[c]{@{}c@{}}task\\ ratio\end{tabular}} 
& \textbf{\begin{tabular}[c]{@{}c@{}}task\\ ratio\end{tabular}}
& \textbf{en-de}
& \textbf{en-it}
& \textbf{en-zh}
& \textbf{\begin{tabular}[c]{@{}c@{}}task\\ ratio\end{tabular}}
& \textbf{en-en}
& \textbf{en-de}
& \textbf{en-it}
& \textbf{en-zh} \\\midrule
Text-only LoRA 
& x                                                             
& 0.6
& 0.4
& 0.3
& 0.3
& 0.4
& 0.2 / 0.05
& 0.2 / 0.05
& 0.2 / 0.05
& 0.2 / 0.05
\\
A.1
& 0.4
& 0.6
& 0.3
& 0.4
& 0.3
& x
& x
& x
& x
& x
\\
A.2
& 0.4
& 0.35
& 0.3
& 0.4
& 0.3
& 0.25
& 0.2 / 0.2
& 0.15 / 0.15
& x
& 0.15 / 0.15
\\
A.1 + B
& 0.2
& 0.4
& 0.4
& 0.3
& 0.3
& 0.4
& 0.2 / 0.05
&0.2 / 0.05
&0.2 / 0.05
&0.2 / 0.05 \\
A.2 + B 
& 0.2 
& 0.4 
& 0.4
& 0.3
& 0.3
& 0.4
& 0.2 / 0.05
&0.2 / 0.05
&0.2 / 0.05
&0.2 / 0.05 \\\bottomrule
\end{tabular}}
\caption{Two-level sampling ratio for each model.}
\label{tab:probabilityinfo}
\end{table*}

%% file: main.bbl
\begin{thebibliography}{34}
\providecommand{\natexlab}[1]{#1}
\providecommand{\url}[1]{\texttt{#1}}
\expandafter\ifx\csname urlstyle\endcsname\relax
  \providecommand{\doi}[1]{doi: #1}\else
  \providecommand{\doi}{doi: \begingroup \urlstyle{rm}\Url}\fi

\bibitem[Abdulmumin et~al.(2025)Abdulmumin, Agostinelli, Alumäe, Anastasopoulos, Ashwin, Bentivogli, Bojar, Borg, Bougares, Cattoni, Cettolo, Chen, Chen, Dabre, Estève, Federico, Gaido, Javorský, Kasztelnik, Lam, Liu, Matusov, Maurya, McCrae, Mdhaffar, Moslem, Murray, Nakamura, Negri, Niehues, Ojha, Ortega, Papi, Pecina, Polák, Połeć, Savoldi, Sethiya, Sikasote, Sperber, Stüker, Sudoh, Thompson, Turchi, Waibel, Wilken, Zevallos, Zouhar, and Züfle]{abdulmumin-etal-2025-findings}
Idris Abdulmumin, Victor Agostinelli, Tanel Alumäe, Antonios Anastasopoulos, Ashwin, Luisa Bentivogli, Ondřej Bojar, Claudia Borg, Fethi Bougares, Roldano Cattoni, Mauro Cettolo, Lizhong Chen, William Chen, Raj Dabre, Yannick Estève, Marcello Federico, Marco Gaido, Dávid Javorský, Marek Kasztelnik, Tsz~Kin Lam, Danni Liu, Evgeny Matusov, Chandresh~Kumar Maurya, John~P. McCrae, Salima Mdhaffar, Yasmin Moslem, Kenton Murray, Satoshi Nakamura, Matteo Negri, Jan Niehues, Atul~Kr. Ojha, John~E. Ortega, Sara Papi, Pavel Pecina, Peter Polák, Piotr Połeć, Beatrice Savoldi, Nivedita Sethiya, Claytone Sikasote, Matthias Sperber, Sebastian Stüker, Katsuhito Sudoh, Brian Thompson, Marco Turchi, Alex Waibel, Patrick Wilken, Rodolfo Zevallos, Vilém Zouhar, and Maike Züfle.
\newblock Findings of the iwslt 2025 evaluation campaign.
\newblock In \emph{Proceedings of the 22nd International Conference on Spoken Language Translation (IWSLT 2025)}, Vienna, Austria (in-person and online), 2025. Association for Computational Linguistics.
\newblock To appear.

\bibitem[Achiam et~al.(2023)Achiam, Adler, Agarwal, Ahmad, Akkaya, Aleman, Almeida, Altenschmidt, Altman, Anadkat, et~al.]{achiam2023gpt}
Josh Achiam, Steven Adler, Sandhini Agarwal, Lama Ahmad, Ilge Akkaya, Florencia~Leoni Aleman, Diogo Almeida, Janko Altenschmidt, Sam Altman, Shyamal Anadkat, et~al.
\newblock Gpt-4 technical report.
\newblock \emph{arXiv preprint arXiv:2303.08774}, 2023.

\bibitem[Alves et~al.(2024)Alves, Pombal, Guerreiro, Martins, Alves, Farajian, Peters, Rei, Fernandes, Agrawal, Colombo, de~Souza, and Martins]{alves2024tower}
Duarte~Miguel Alves, Jos{\'e} Pombal, Nuno~M Guerreiro, Pedro~Henrique Martins, Jo{\~a}o Alves, Amin Farajian, Ben Peters, Ricardo Rei, Patrick Fernandes, Sweta Agrawal, Pierre Colombo, Jos{\'e} G.~C. de Souza, and Andre Martins.
\newblock Tower: An open multilingual large language model for translation-related tasks.
\newblock In \emph{First Conference on Language Modeling}, 2024.

\bibitem[Ambilduke et~al.(2025)Ambilduke, Peters, Sannigrahi, Keshwani, Lam, Martins, Boito, and Martins]{ambilduke2025tower}
Kshitij Ambilduke, Ben Peters, Sonal Sannigrahi, Anil Keshwani, Tsz~Kin Lam, Bruno Martins, Marcely~Zanon Boito, and Andr{\'e}~FT Martins.
\newblock From tower to spire: Adding the speech modality to a text-only llm.
\newblock \emph{arXiv preprint arXiv:2503.10620}, 2025.

\bibitem[Barrault et~al.(2023)Barrault, Chung, Meglioli, Dale, Dong, Duquenne, Elsahar, Gong, Heffernan, Hoffman, et~al.]{barrault2023seamlessm4t}
Lo{\"\i}c Barrault, Yu-An Chung, Mariano~Cora Meglioli, David Dale, Ning Dong, Paul-Ambroise Duquenne, Hady Elsahar, Hongyu Gong, Kevin Heffernan, John Hoffman, et~al.
\newblock Seamlessm4t: Massively multilingual \& multimodal machine translation.
\newblock \emph{arXiv preprint arXiv:2308.11596}, 2023.

\bibitem[D{\'e}fossez et~al.(2024)D{\'e}fossez, Mazar{\'e}, Orsini, Royer, P{\'e}rez, J{\'e}gou, Grave, and Zeghidour]{defossez2024moshi}
Alexandre D{\'e}fossez, Laurent Mazar{\'e}, Manu Orsini, Am{\'e}lie Royer, Patrick P{\'e}rez, Herv{\'e} J{\'e}gou, Edouard Grave, and Neil Zeghidour.
\newblock Moshi: a speech-text foundation model for real-time dialogue.
\newblock \emph{arXiv preprint arXiv:2410.00037}, 2024.

\bibitem[Driess et~al.(2023)Driess, Xia, Sajjadi, Lynch, Chowdhery, Ichter, Wahid, Tompson, Vuong, Yu, et~al.]{driess2023palm}
Danny Driess, Fei Xia, Mehdi~SM Sajjadi, Corey Lynch, Aakanksha Chowdhery, Brian Ichter, Ayzaan Wahid, Jonathan Tompson, Quan Vuong, Tianhe Yu, et~al.
\newblock Palm-e: An embodied multimodal language model.
\newblock \emph{arXiv preprint arXiv:2303.03378}, 2023.

\bibitem[Grattafiori et~al.(2024)Grattafiori, Dubey, Jauhri, Pandey, Kadian, Al-Dahle, Letman, Mathur, Schelten, Vaughan, et~al.]{grattafiori2024llama}
Aaron Grattafiori, Abhimanyu Dubey, Abhinav Jauhri, Abhinav Pandey, Abhishek Kadian, Ahmad Al-Dahle, Aiesha Letman, Akhil Mathur, Alan Schelten, Alex Vaughan, et~al.
\newblock The llama 3 herd of models.
\newblock \emph{arXiv preprint arXiv:2407.21783}, 2024.

\bibitem[Hu et~al.(2022)Hu, Shen, Wallis, Allen-Zhu, Li, Wang, Wang, and Chen]{hu2022lora}
Edward~J Hu, Yelong Shen, Phillip Wallis, Zeyuan Allen-Zhu, Yuanzhi Li, Shean Wang, Lu Wang, and Weizhu Chen.
\newblock Lo{RA}: Low-rank adaptation of large language models.
\newblock In \emph{International Conference on Learning Representations}, 2022.

\bibitem[Hu et~al.(2024)Hu, Zhou, Liu, Chen, Meng, Hao, Pan, Liu, Li, Sivasankaran, et~al.]{hu2024wavllm}
Shujie Hu, Long Zhou, Shujie Liu, Sanyuan Chen, Lingwei Meng, Hongkun Hao, Jing Pan, Xunying Liu, Jinyu Li, Sunit Sivasankaran, et~al.
\newblock Wavllm: Towards robust and adaptive speech large language model.
\newblock \emph{arXiv preprint arXiv:2404.00656}, 2024.

\bibitem[Huang et~al.(2024)Huang, Li, Yang, Shi, Chang, Ye, Wu, Hong, Huang, Liu, et~al.]{huang2024audiogpt}
Rongjie Huang, Mingze Li, Dongchao Yang, Jiatong Shi, Xuankai Chang, Zhenhui Ye, Yuning Wu, Zhiqing Hong, Jiawei Huang, Jinglin Liu, et~al.
\newblock Audiogpt: Understanding and generating speech, music, sound, and talking head.
\newblock In \emph{Proceedings of the AAAI Conference on Artificial Intelligence}, pages 23802--23804, 2024.

\bibitem[{Iranzo-Sánchez} et~al.(2020){Iranzo-Sánchez}, {Silvestre-Cerdà}, {Jorge}, {Roselló}, {Giménez}, {Sanchis}, {Civera}, and {Juan}]{jairsan2020a}
J. {Iranzo-Sánchez}, J.~A. {Silvestre-Cerdà}, J. {Jorge}, N. {Roselló}, A. {Giménez}, A. {Sanchis}, J. {Civera}, and A. {Juan}.
\newblock Europarl-st: A multilingual corpus for speech translation of parliamentary debates.
\newblock In \emph{ICASSP 2020 - 2020 IEEE International Conference on Acoustics, Speech and Signal Processing (ICASSP)}, pages 8229--8233, 2020.

\bibitem[Jiang et~al.(2024)Jiang, Sablayrolles, Roux, Mensch, Savary, Bamford, Chaplot, Casas, Hanna, Bressand, et~al.]{jiang2024mixtral}
Albert~Q Jiang, Alexandre Sablayrolles, Antoine Roux, Arthur Mensch, Blanche Savary, Chris Bamford, Devendra~Singh Chaplot, Diego de~las Casas, Emma~Bou Hanna, Florian Bressand, et~al.
\newblock Mixtral of experts.
\newblock \emph{arXiv preprint arXiv:2401.04088}, 2024.

\bibitem[Laurençon et~al.(2024)Laurençon, Tronchon, Cord, and Sanh]{laurencon_what_2024}
Hugo Laurençon, Léo Tronchon, Matthieu Cord, and Victor Sanh.
\newblock What matters when building vision-language models?, 2024.
\newblock arXiv:2405.02246.

\bibitem[Lee et~al.(2018)Lee, Wu, Liu, and Lee]{lee2018spoken}
Chia-Hsuan Lee, Szu-Lin Wu, Chi-Liang Liu, and Hung-yi Lee.
\newblock Spoken squad: A study of mitigating the impact of speech recognition errors on listening comprehension.
\newblock \emph{Proc. Interspeech 2018}, pages 3459--3463, 2018.

\bibitem[Lin(2004)]{lin-2004-rouge}
Chin-Yew Lin.
\newblock {ROUGE}: A package for automatic evaluation of summaries.
\newblock In \emph{Text Summarization Branches Out}, pages 74--81, Barcelona, Spain, 2004. Association for Computational Linguistics.

\bibitem[Liu et~al.(2023)Liu, Li, Wu, and Lee]{liu_visual_2023}
Haotian Liu, Chunyuan Li, Qingyang Wu, and Yong~Jae Lee.
\newblock Visual {Instruction} {Tuning} ({LLaVA}), 2023.
\newblock arXiv:2304.08485 [cs].

\bibitem[Martins et~al.(2024{\natexlab{a}})Martins, Fernandes, Alves, Guerreiro, Rei, Alves, Pombal, Farajian, Faysse, Klimaszewski, Colombo, Haddow, de~Souza, Birch, and Martins]{eurollm}
Pedro~Henrique Martins, Patrick Fernandes, João Alves, Nuno~M. Guerreiro, Ricardo Rei, Duarte~M. Alves, José Pombal, Amin Farajian, Manuel Faysse, Mateusz Klimaszewski, Pierre Colombo, Barry Haddow, José G.~C. de Souza, Alexandra Birch, and André F.~T. Martins.
\newblock Eurollm: Multilingual language models for europe, 2024{\natexlab{a}}.

\bibitem[Martins et~al.(2024{\natexlab{b}})Martins, Fernandes, Alves, Guerreiro, Rei, Alves, Pombal, Farajian, Faysse, Klimaszewski, et~al.]{martins2024eurollm}
Pedro~Henrique Martins, Patrick Fernandes, Jo{\~a}o Alves, Nuno~M Guerreiro, Ricardo Rei, Duarte~M Alves, Jos{\'e} Pombal, Amin Farajian, Manuel Faysse, Mateusz Klimaszewski, et~al.
\newblock Eurollm: Multilingual language models for europe.
\newblock \emph{arXiv preprint arXiv:2409.16235}, 2024{\natexlab{b}}.

\bibitem[Nguyen et~al.(2025)Nguyen, Muller, Yu, Costa-Jussa, Elbayad, Popuri, Ropers, Duquenne, Algayres, Mavlyutov, et~al.]{nguyen2024spirit}
Tu~Anh Nguyen, Benjamin Muller, Bokai Yu, Marta~R Costa-Jussa, Maha Elbayad, Sravya Popuri, Christophe Ropers, Paul-Ambroise Duquenne, Robin Algayres, Ruslan Mavlyutov, et~al.
\newblock Spirit-lm: Interleaved spoken and written language model.
\newblock \emph{Transactions of the Association for Computational Linguistics}, 13:\penalty0 30--52, 2025.

\bibitem[Pikabea et~al.(2025)Pikabea, Lacunza, Pareras, Escolano, Gonzalez-Agirre, Hernando, and Villegas]{pikabea2025breaking}
I{\~n}igo Pikabea, I{\~n}aki Lacunza, Oriol Pareras, Carlos Escolano, Aitor Gonzalez-Agirre, Javier Hernando, and Marta Villegas.
\newblock Breaking language barriers in visual language models via multilingual textual regularization.
\newblock \emph{arXiv preprint arXiv:2503.22577}, 2025.

\bibitem[Post(2018)]{post-2018-call}
Matt Post.
\newblock A call for clarity in reporting {BLEU} scores.
\newblock In \emph{Proceedings of the Third Conference on Machine Translation: Research Papers}, pages 186--191, Belgium, Brussels, 2018. Association for Computational Linguistics.

\bibitem[Pratap et~al.(2024)Pratap, Tjandra, Shi, Tomasello, Babu, Kundu, Elkahky, Ni, Vyas, Fazel-Zarandi, et~al.]{pratap2024scaling}
Vineel Pratap, Andros Tjandra, Bowen Shi, Paden Tomasello, Arun Babu, Sayani Kundu, Ali Elkahky, Zhaoheng Ni, Apoorv Vyas, Maryam Fazel-Zarandi, et~al.
\newblock Scaling speech technology to 1,000+ languages.
\newblock \emph{Journal of Machine Learning Research}, 25\penalty0 (97):\penalty0 1--52, 2024.

\bibitem[Rau et~al.(2024)Rau, D{\'e}jean, Chirkova, Formal, Wang, Clinchant, and Nikoulina]{rau-etal-2024-bergen}
David Rau, Herv{\'e} D{\'e}jean, Nadezhda Chirkova, Thibault Formal, Shuai Wang, St{\'e}phane Clinchant, and Vassilina Nikoulina.
\newblock {BERGEN}: A benchmarking library for retrieval-augmented generation.
\newblock In \emph{Findings of the Association for Computational Linguistics: EMNLP 2024}, pages 7640--7663, Miami, Florida, USA, 2024. Association for Computational Linguistics.

\bibitem[Rei et~al.(2022)Rei, C.~de Souza, Alves, Zerva, Farinha, Glushkova, Lavie, Coheur, and Martins]{rei-etal-2022-comet}
Ricardo Rei, Jos{\'e}~G. C.~de Souza, Duarte Alves, Chrysoula Zerva, Ana~C Farinha, Taisiya Glushkova, Alon Lavie, Luisa Coheur, and Andr{\'e} F.~T. Martins.
\newblock {COMET}-22: Unbabel-{IST} 2022 submission for the metrics shared task.
\newblock In \emph{Proceedings of the Seventh Conference on Machine Translation (WMT)}, pages 578--585, Abu Dhabi, United Arab Emirates (Hybrid), 2022. Association for Computational Linguistics.

\bibitem[Rubenstein et~al.(2023)Rubenstein, Asawaroengchai, Nguyen, Bapna, Borsos, Quitry, Chen, Badawy, Han, Kharitonov, et~al.]{rubenstein2023audiopalm}
Paul~K Rubenstein, Chulayuth Asawaroengchai, Duc~Dung Nguyen, Ankur Bapna, Zal{\'a}n Borsos, F{\'e}lix de~Chaumont Quitry, Peter Chen, Dalia~El Badawy, Wei Han, Eugene Kharitonov, et~al.
\newblock Audiopalm: A large language model that can speak and listen.
\newblock \emph{arXiv preprint arXiv:2306.12925}, 2023.

\bibitem[Salesky et~al.(2023)Salesky, Darwish, Al-Badrashiny, Diab, and Niehues]{salesky-etal-2023-evaluating}
Elizabeth Salesky, Kareem Darwish, Mohamed Al-Badrashiny, Mona Diab, and Jan Niehues.
\newblock Evaluating multilingual speech translation under realistic conditions with resegmentation and terminology.
\newblock In \emph{Proceedings of the 20th International Conference on Spoken Language Translation (IWSLT 2023)}, pages 62--78, Toronto, Canada (in-person and online), 2023. Association for Computational Linguistics.

\bibitem[Tang et~al.(2023)Tang, Yu, Sun, Chen, Tan, Li, Lu, Ma, and Zhang]{tang2023salmonn}
Changli Tang, Wenyi Yu, Guangzhi Sun, Xianzhao Chen, Tian Tan, Wei Li, Lu Lu, Zejun Ma, and Chao Zhang.
\newblock Salmonn: Towards generic hearing abilities for large language models.
\newblock \emph{arXiv preprint arXiv:2310.13289}, 2023.

\bibitem[Team et~al.(2023)Team, Anil, Borgeaud, Alayrac, Yu, Soricut, Schalkwyk, Dai, Hauth, Millican, et~al.]{team2023gemini}
Gemini Team, Rohan Anil, Sebastian Borgeaud, Jean-Baptiste Alayrac, Jiahui Yu, Radu Soricut, Johan Schalkwyk, Andrew~M Dai, Anja Hauth, Katie Millican, et~al.
\newblock Gemini: a family of highly capable multimodal models.
\newblock \emph{arXiv preprint arXiv:2312.11805}, 2023.

\bibitem[Team et~al.(2025)Team, Kamath, Ferret, Pathak, Vieillard, Merhej, Perrin, Matejovicova, Ramé, Rivière, Rouillard, Mesnard, Cideron, bastien Grill, Ramos, Yvinec, Casbon, Pot, Penchev, Liu, Visin, Kenealy, Beyer, Zhai, Tsitsulin, Busa-Fekete, Feng, Sachdeva, Coleman, Gao, Mustafa, Barr, Parisotto, Tian, Eyal, Cherry, Peter, Sinopalnikov, Bhupatiraju, Agarwal, Kazemi, Malkin, Kumar, Vilar, Brusilovsky, Luo, Steiner, Friesen, Sharma, Sharma, Gilady, Goedeckemeyer, Saade, Feng, Kolesnikov, Bendebury, Abdagic, Vadi, György, Pinto, Das, Bapna, Miech, Yang, Paterson, Shenoy, Chakrabarti, Piot, Wu, Shahriari, Petrini, Chen, Lan, Choquette-Choo, Carey, Brick, Deutsch, Eisenbud, Cattle, Cheng, Paparas, Sreepathihalli, Reid, Tran, Zelle, Noland, Huizenga, Kharitonov, Liu, Amirkhanyan, Cameron, Hashemi, Klimczak-Plucińska, Singh, Mehta, Lehri, Hazimeh, Ballantyne, Szpektor, Nardini, Pouget-Abadie, Chan, Stanton, Wieting, Lai, Orbay, Fernandez, Newlan, yeong Ji, Singh, Black, Yu, Hui, Vodrahalli, Greff, Qiu,
  Valentine, Coelho, Ritter, Hoffman, Watson, Chaturvedi, Moynihan, Ma, Babar, Noy, Byrd, Roy, Momchev, Chauhan, Sachdeva, Bunyan, Botarda, Caron, Rubenstein, Culliton, Schmid, Sessa, Xu, Stanczyk, Tafti, Shivanna, Wu, Pan, Rokni, Willoughby, Vallu, Mullins, Jerome, Smoot, Girgin, Iqbal, Reddy, Sheth, Põder, Bhatnagar, Panyam, Eiger, Zhang, Liu, Yacovone, Liechty, Kalra, Evci, Misra, Roseberry, Feinberg, Kolesnikov, Han, Kwon, Chen, Chow, Zhu, Wei, Egyed, Cotruta, Giang, Kirk, Rao, Black, Babar, Lo, Moreira, Martins, Sanseviero, Gonzalez, Gleicher, Warkentin, Mirrokni, Senter, Collins, Barral, Ghahramani, Hadsell, Matias, Sculley, Petrov, Fiedel, Shazeer, Vinyals, Dean, Hassabis, Kavukcuoglu, Farabet, Buchatskaya, Alayrac, Anil, Dmitry, Lepikhin, Borgeaud, Bachem, Joulin, Andreev, Hardin, Dadashi, and Hussenot]{gemma3}
Gemma Team, Aishwarya Kamath, Johan Ferret, Shreya Pathak, Nino Vieillard, Ramona Merhej, Sarah Perrin, Tatiana Matejovicova, Alexandre Ramé, Morgane Rivière, Louis Rouillard, Thomas Mesnard, Geoffrey Cideron, Jean bastien Grill, Sabela Ramos, Edouard Yvinec, Michelle Casbon, Etienne Pot, Ivo Penchev, Gaël Liu, Francesco Visin, Kathleen Kenealy, Lucas Beyer, Xiaohai Zhai, Anton Tsitsulin, Robert Busa-Fekete, Alex Feng, Noveen Sachdeva, Benjamin Coleman, Yi Gao, Basil Mustafa, Iain Barr, Emilio Parisotto, David Tian, Matan Eyal, Colin Cherry, Jan-Thorsten Peter, Danila Sinopalnikov, Surya Bhupatiraju, Rishabh Agarwal, Mehran Kazemi, Dan Malkin, Ravin Kumar, David Vilar, Idan Brusilovsky, Jiaming Luo, Andreas Steiner, Abe Friesen, Abhanshu Sharma, Abheesht Sharma, Adi~Mayrav Gilady, Adrian Goedeckemeyer, Alaa Saade, Alex Feng, Alexander Kolesnikov, Alexei Bendebury, Alvin Abdagic, Amit Vadi, András György, André~Susano Pinto, Anil Das, Ankur Bapna, Antoine Miech, Antoine Yang, Antonia Paterson, Ashish
  Shenoy, Ayan Chakrabarti, Bilal Piot, Bo Wu, Bobak Shahriari, Bryce Petrini, Charlie Chen, Charline~Le Lan, Christopher~A. Choquette-Choo, CJ Carey, Cormac Brick, Daniel Deutsch, Danielle Eisenbud, Dee Cattle, Derek Cheng, Dimitris Paparas, Divyashree~Shivakumar Sreepathihalli, Doug Reid, Dustin Tran, Dustin Zelle, Eric Noland, Erwin Huizenga, Eugene Kharitonov, Frederick Liu, Gagik Amirkhanyan, Glenn Cameron, Hadi Hashemi, Hanna Klimczak-Plucińska, Harman Singh, Harsh Mehta, Harshal~Tushar Lehri, Hussein Hazimeh, Ian Ballantyne, Idan Szpektor, Ivan Nardini, Jean Pouget-Abadie, Jetha Chan, Joe Stanton, John Wieting, Jonathan Lai, Jordi Orbay, Joseph Fernandez, Josh Newlan, Ju yeong Ji, Jyotinder Singh, Kat Black, Kathy Yu, Kevin Hui, Kiran Vodrahalli, Klaus Greff, Linhai Qiu, Marcella Valentine, Marina Coelho, Marvin Ritter, Matt Hoffman, Matthew Watson, Mayank Chaturvedi, Michael Moynihan, Min Ma, Nabila Babar, Natasha Noy, Nathan Byrd, Nick Roy, Nikola Momchev, Nilay Chauhan, Noveen Sachdeva, Oskar
  Bunyan, Pankil Botarda, Paul Caron, Paul~Kishan Rubenstein, Phil Culliton, Philipp Schmid, Pier~Giuseppe Sessa, Pingmei Xu, Piotr Stanczyk, Pouya Tafti, Rakesh Shivanna, Renjie Wu, Renke Pan, Reza Rokni, Rob Willoughby, Rohith Vallu, Ryan Mullins, Sammy Jerome, Sara Smoot, Sertan Girgin, Shariq Iqbal, Shashir Reddy, Shruti Sheth, Siim Põder, Sijal Bhatnagar, Sindhu~Raghuram Panyam, Sivan Eiger, Susan Zhang, Tianqi Liu, Trevor Yacovone, Tyler Liechty, Uday Kalra, Utku Evci, Vedant Misra, Vincent Roseberry, Vlad Feinberg, Vlad Kolesnikov, Woohyun Han, Woosuk Kwon, Xi Chen, Yinlam Chow, Yuvein Zhu, Zichuan Wei, Zoltan Egyed, Victor Cotruta, Minh Giang, Phoebe Kirk, Anand Rao, Kat Black, Nabila Babar, Jessica Lo, Erica Moreira, Luiz~Gustavo Martins, Omar Sanseviero, Lucas Gonzalez, Zach Gleicher, Tris Warkentin, Vahab Mirrokni, Evan Senter, Eli Collins, Joelle Barral, Zoubin Ghahramani, Raia Hadsell, Yossi Matias, D. Sculley, Slav Petrov, Noah Fiedel, Noam Shazeer, Oriol Vinyals, Jeff Dean, Demis Hassabis,
  Koray Kavukcuoglu, Clement Farabet, Elena Buchatskaya, Jean-Baptiste Alayrac, Rohan Anil, Dmitry, Lepikhin, Sebastian Borgeaud, Olivier Bachem, Armand Joulin, Alek Andreev, Cassidy Hardin, Robert Dadashi, and Léonard Hussenot.
\newblock Gemma 3 technical report, 2025.

\bibitem[torchtune maintainers and contributors(2024)]{torchtune}
torchtune maintainers and contributors.
\newblock torchtune: Pytorch's finetuning library, 2024.

\bibitem[Touvron et~al.(2023)Touvron, Martin, Stone, Albert, Almahairi, Babaei, Bashlykov, Batra, Bhargava, Bhosale, Bikel, Blecher, Ferrer, Chen, Cucurull, Esiobu, Fernandes, Fu, Fu, Fuller, Gao, Goswami, Goyal, Hartshorn, Hosseini, Hou, Inan, Kardas, Kerkez, Khabsa, Kloumann, Korenev, Koura, Lachaux, Lavril, Lee, Liskovich, Lu, Mao, Martinet, Mihaylov, Mishra, Molybog, Nie, Poulton, Reizenstein, Rungta, Saladi, Schelten, Silva, Smith, Subramanian, Tan, Tang, Taylor, Williams, Kuan, Xu, Yan, Zarov, Zhang, Fan, Kambadur, Narang, Rodriguez, Stojnic, Edunov, and Scialom]{touvron2023llama}
Hugo Touvron, Louis Martin, Kevin Stone, Peter Albert, Amjad Almahairi, Yasmine Babaei, Nikolay Bashlykov, Soumya Batra, Prajjwal Bhargava, Shruti Bhosale, Dan Bikel, Lukas Blecher, Cristian~Canton Ferrer, Moya Chen, Guillem Cucurull, David Esiobu, Jude Fernandes, Jeremy Fu, Wenyin Fu, Brian Fuller, Cynthia Gao, Vedanuj Goswami, Naman Goyal, Anthony Hartshorn, Saghar Hosseini, Rui Hou, Hakan Inan, Marcin Kardas, Viktor Kerkez, Madian Khabsa, Isabel Kloumann, Artem Korenev, Punit~Singh Koura, Marie-Anne Lachaux, Thibaut Lavril, Jenya Lee, Diana Liskovich, Yinghai Lu, Yuning Mao, Xavier Martinet, Todor Mihaylov, Pushkar Mishra, Igor Molybog, Yixin Nie, Andrew Poulton, Jeremy Reizenstein, Rashi Rungta, Kalyan Saladi, Alan Schelten, Ruan Silva, Eric~Michael Smith, Ranjan Subramanian, Xiaoqing~Ellen Tan, Binh Tang, Ross Taylor, Adina Williams, Jian~Xiang Kuan, Puxin Xu, Zheng Yan, Iliyan Zarov, Yuchen Zhang, Angela Fan, Melanie Kambadur, Sharan Narang, Aurelien Rodriguez, Robert Stojnic, Sergey Edunov, and Thomas
  Scialom.
\newblock Llama 2: Open foundation and fine-tuned chat models, 2023.

\bibitem[Wang et~al.(2020)Wang, Wu, and Pino]{wang2020covost}
Changhan Wang, Anne Wu, and Juan Pino.
\newblock Covost 2: A massively multilingual speech-to-text translation corpus, 2020.

\bibitem[Yang et~al.(2024)Yang, Yang, Zhang, Hui, Zheng, Yu, Li, Liu, Huang, Wei, et~al.]{yang2024qwen2}
An Yang, Baosong Yang, Beichen Zhang, Binyuan Hui, Bo Zheng, Bowen Yu, Chengyuan Li, Dayiheng Liu, Fei Huang, Haoran Wei, et~al.
\newblock Qwen2. 5 technical report.
\newblock \emph{arXiv preprint arXiv:2412.15115}, 2024.

\end{thebibliography}
